\RequirePackage{fix-cm}
\documentclass{svjour3}                     
\smartqed  
\usepackage{graphicx}
\usepackage{amsmath,epsfig}
\usepackage{verbatim}
\usepackage{multirow}
\usepackage{balance}

\usepackage[numbers,sort&compress]{natbib}
\setcitestyle{square}

\usepackage{amsfonts}
\usepackage[algoruled,boxed,lined]{algorithm2e}
\makeatother
\SetKwProg{Fn}{Function}{}{}
\DeclareMathOperator*{\argmax}{argmax}

\newcommand{\BState}[0]{\State\hskip-\ALG@thistlm}
\journalname{Multimedia Tools and Applications}
\begin{document}

\title{Dynamic Character Graph via Online Face Clustering for Movie Analysis}

\author{Prakhar Kulshreshtha         \and
        Tanaya Guha 
}


\institute{P. Kulshreshtha \at
              Carnegie Mellon University, Pittsburgh, US \\
              \email{pkulshre@andrew.cmu.edu}           
           \and
           T. Guha \at
           University of Warwick, Coventry, UK\\
           \email{tanaya.guha@warwick.ac.uk}
}


\maketitle

\begin{abstract}
An effective approach to automated movie content analysis involves building a network (graph) of its characters. Existing work usually builds a static character graph to summarize the content using metadata, scripts or manual annotations. We propose an unsupervised approach to building a \emph{dynamic} character graph that captures the temporal evolution of character interaction. We refer to this as the \emph{character interaction graph} (CIG). Our approach has two components: (i) an online face clustering algorithm that discovers the characters in the video stream as they appear, and (ii) simultaneous creation of a CIG using the temporal dynamics of the resulting clusters. We demonstrate the usefulness of the CIG for two movie analysis tasks: narrative structure (acts) segmentation, and major character retrieval. Our evaluation on full-length movies containing more than 5000 face tracks shows that the proposed approach achieves superior performance for both the tasks.

\end{abstract}

\section{Introduction}
\label{sec:intro}
Automated analysis of media content, such as movies has traditionally focused on extracting and using low level features from shots and scenes for analyzing narrative structures and key events \cite{li2004content,li2006techniques}. For humans, however, a movie is not just a collection of shots or scenes. It is the characters that usually play the most important role in storytelling \cite{sharff1982elements}.
More recently, character-centric representation of movies, such as character networks have emerged as an effective approach towards media content analysis \cite{rolenet, characternet, ramakrishna2017linguistic}. A character network usually has the major characters as its nodes where the edges summarize the relationship between character pairs. Such networks have been shown to facilitate a number of movie analysis tasks including character analysis \cite{ramakrishna2017linguistic}, story segmentation \cite{rolenet} and major character identification \cite{characternet}.
The existing methods build a single, static character network for the entire movie. While static graphs offer a convenient summary of the overall interactions among characters, they can not capture the evolution of a movie's dynamic narrative. 

\begin{figure}[tb]
    \centering
  \includegraphics[width=1.0\linewidth, trim= {0cm 0cm 0cm 0cm}, clip]{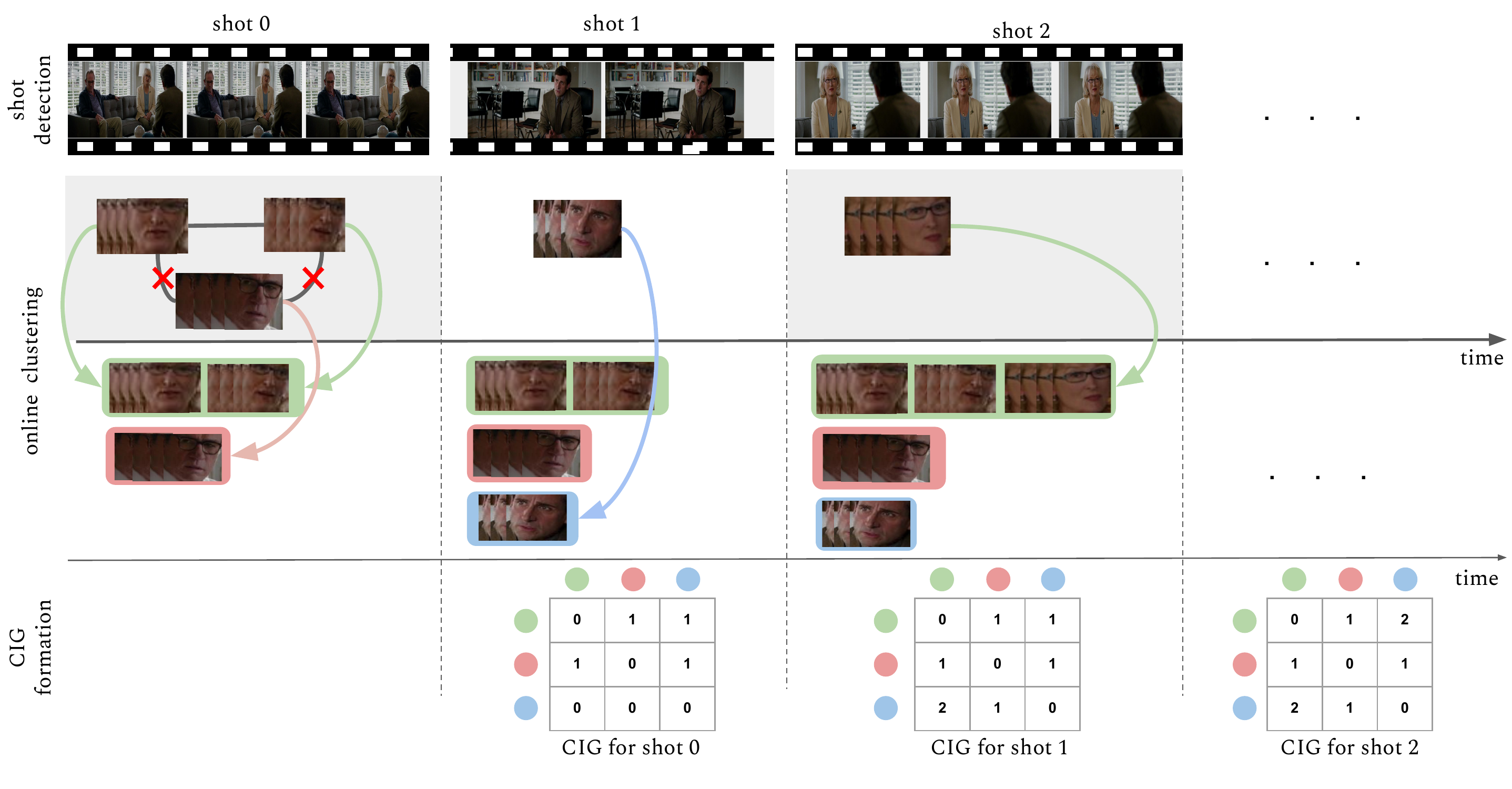}
	\caption{Overview of the proposed approach: A movie is processed at shot level. For each shot, face tracks are created, and our online clustering algorithm either group a face track with an existing cluster or create a new one. In this example, at shot0, 3 face tracks are created and grouped into two clusters. A new cluster is added in shot1 as it belongs to a new character. The face track in shot2 is added to an existing cluster belonging to the same character. The CIG is updated after each shot is processing. Note that the CIG for the $(i-1)^{th}$ shot is obtained after the $(i-1)^{th}$ shot is processed.}
	\label{fig:OCC}
\end{figure}

In this paper, we present an unsupervised approach to building a \emph{dynamic} character network via online face clustering. We refer to this network as the \emph{character interaction graph} (CIG), where each movie character is represented as a node and an edge represents pairwise interaction between characters. The dynamic aspect of the CIG offers an effective way to capture the variations in character interactions over time - particularly helpful for story segmentation and event localization. Our approach (see Fig.~\ref{fig:OCC} for an overview) has of two main components - online face clustering, and simultaneous creation of the CIG using the resulting clusters. Building on our previous work on online face clustering \cite{kulshreshtha2018online}, we 
develop a new algorithm to create (and update) a CIG via clustering i.e.,\ utilizing the information from the cluster dynamics. We demonstrate the usefulness of the CIGs for two important movie analysis tasks: (i) semantic segmentation of a movie into \emph{acts}, and (ii) major character discovery. Performance is evaluated on a database of six full-length Hollywood movies containing more than 5000 face tracks. Results are compared with relevant past work and manual annotations, where our CIG-based approach shows superior performance. 

In summary, the contribution of this work is two-fold: (i) We propose an unsupervised approach to building dynamic character graph via online face clustering. This is the first work on dynamic CIG construction. (ii) We demonstrate superior performance of our CIG-based approach for two important movie analysis tasks - three act segmentation and major character identification.

The rest of this paper is organized as follows. Section \ref{sec:rel_work} discusses relevant literature for character network-based movie analysis and online face clustering. Section \ref{sec:method} describes our approach to dynamic CIG creation via online face clustering. Section \ref{sec:application} proposes the methodologies to apply CIG for two movie analysis tasks. Section \ref{sec:evaluation} presents extensive results, and Section \ref{sec:conclusion} concludes the article with summary and discussion on future work.%
\section{Related work}
\label{sec:rel_work}
Following the two major components of our approach, we discuss past work related to character graph construction and face clustering in multimedia content.
\subsection{Character network construction}
Character networks are useful for multimedia content analysis due to its wide applicability in story summarization, segmentation, character identification, and character-based search and indexing \citep{rolenet, characternet, ramakrishna2017linguistic, cocharnet}. Character networks have been constructed using movie scripts \cite{ramakrishna2017linguistic}, spoken dialogs \cite{characternet}, manually-labeled data \cite{rolenet}, and supervised face recognition \cite{rolenet}. 

Ramakrishna et al. \citep{ramakrishna2017linguistic} used scripts to construct a character network, where an edge between two characters (nodes) is added if the characters have consecutive dialogs. This network is used to examine the character analytics based on gender, race and age \citep{ramakrishna2017linguistic}. Weng et al. \cite{rolenet} constructed a character network, called the RoleNet, that captures the co-occurrence statistics of movie characters via face recognition. This network is used to identify the lead characters and communities, and for story segmentation \cite{rolenet}. Park et al. \cite{characternet} built a network aligning scripts and subtitles. This network is employed in classification of major and minor characters, community clustering and sequence detection \cite{characternet}. Along the similar lines, Tran and Jung constructed a CoCharNet \cite{cocharnet} using manual annotations to encode information regarding character co-occurrences.

The work most related to our work is that of Yeh and Wu \cite{yeh2014clustering}, where character network is constructed using face clustering. This work clusters faces and construct a character network in an iterative fashion. However, this requires prior knowledge of the number of clusters, and is an offline method. To the best of our knowledge, this is the only prior work that uses (offline) face clustering for constructing character graphs.
\subsection{Face clustering in videos}
\textbf{Offline methods.} The problem of unsupervised face clustering is relatively less studied as compared to its supervised counterpart, i.e., face recognition. The dominant approach to face clustering involves learning a suitable distance measure between face pairs \cite{facenet, jointface,sun2014deep, le2013building}. Several methods have proposed to use partial supervision to improve performance \cite{du2012face, wolf2011face}. While image-based clustering is more common, face clustering in videos can achieve significant improvement by exploiting the temporal information about the faces \cite{constrainedclustering, constrainedmvclustering, simultaneousclustering, weightedblock}. 
Temporal constraints have been used in frameworks based on hidden Markov random field (HMRF) \cite{constrainedclustering} and unsupervised logistic discriminative metric learning (ULDML) \cite{uldml} with applications to face clustering in movies and TV series. A constrained multiview face clustering technique used constrained sparse subspace representation of faces with constrained spectral clustering \cite{constrainedmvclustering}.  
Recent clustering approaches use convolutional neural networks (CNN) to learn robust face representations by using aggregated deep features \cite{simplefc}, deep features with pairwise constraints \cite{jointface} and deep features with triplet loss \cite{tripletloss}. 

\noindent\textbf{Online methods.} The approaches discussed above are all \emph{offline} methods i.e., they assume the availability of the entire data at once. In an online setting, a clustering algorithm does not have the luxury of `seeing' the entire data simultaneously. 
To the best of our knowledge, there is only one work on online face clustering in videos in the existing literature \cite{bayesianentity}. This work created small tracklets of faces from the video, and clustered them in an online fashion based on temporal coherence and the Chinese restaurant process (TCCRP) \cite{bayesianentity}. An extension of this work is Temporally Coherent Chinese Restaurant Franchise (TCCRF) \cite{mitra2014temporally}, that jointly model short
temporal segments. These online methods tend to create multiple clusters for the same person thereby degrading the completeness of the clusters \cite{bayesianentity}. 
\section{Proposed approach}
\label{sec:method}
\textbf{Overview.} In our dynamic CIG construction approach, we process a movie stream at short-level, where a shot is a contiguously recorded sequence of frames. Our approach consists of two main components: (i) face track creation and clustering, and (ii) CIG formation and update. All the components are executed simultaneously in an \emph{online} fashion by processing one shot at a time. As a shot appears, all faces are detected frame by frame and face tracks are created. Our online clustering algorithm then assigns the face tracks to either an existing cluster or to a new one. The information about the cluster updates including formation of new clusters are used to create a dynamic CIG. Fig.~\ref{fig:OCC} presents an overview of the proposed method. Below, we describe each component in detail.
\begin{figure}[tb]
\centering
\includegraphics[width=0.4\linewidth, trim = {0cm 9cm 0cm 0.5cm}, clip]{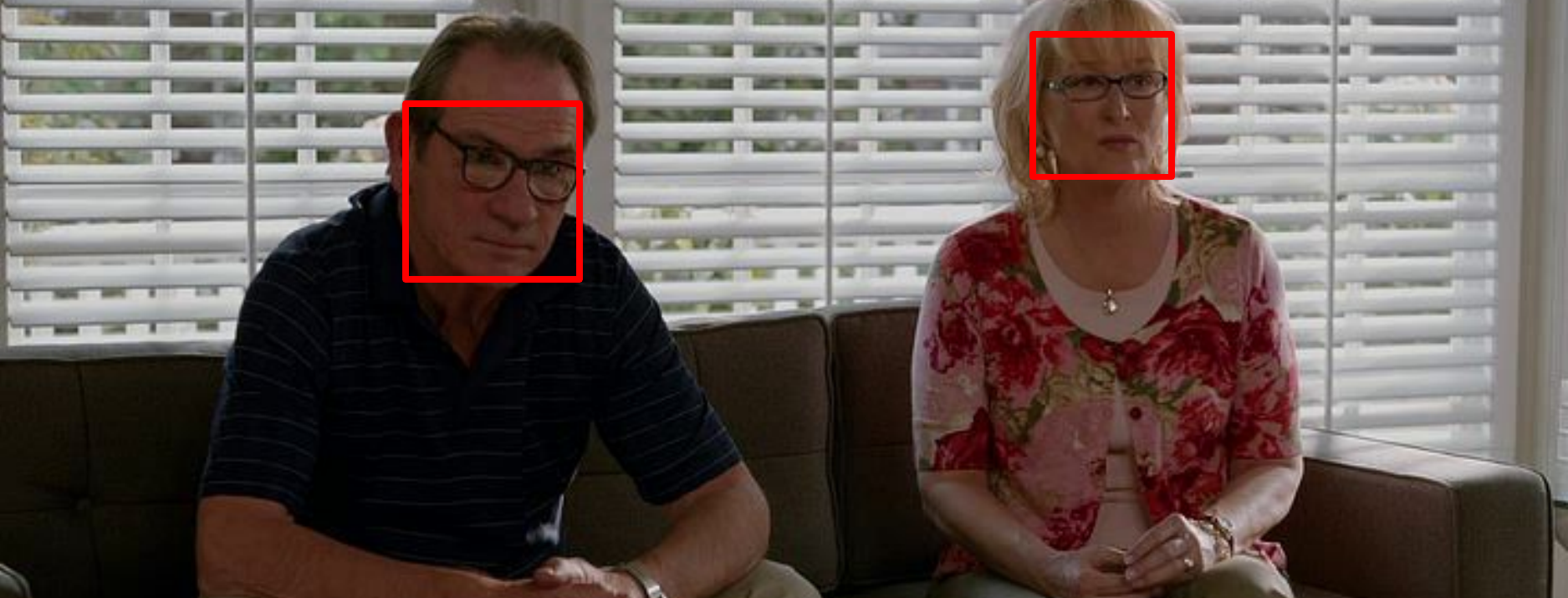}
\includegraphics[width=0.4\linewidth, trim = {0cm 0.5cm 0cm 9cm}, clip]{images/Track_fig_HS.pdf}
\caption{Example of 85\% spatial overlap between face pairs in two consecutive frames which are combined to create face tracks.}
\label{fig:FT}
\end{figure}
\subsection{Face track creation and clustering}

 \textbf{Face track creation.} Consider a movie $\mathcal{M}$ comprising $T$ frames: $\mathcal{M}=\{I_t\}_{t=1}^T$. We define the $i^{th}$ shot $S_i$ as a sequence of consecutive frames $\{I_{t_{(i-1)} + 1}, \ldots I_{t_i}\}$, where $t_i$ is the $i^{th}$ shot boundary. The shot boundary $t_i$ corresponding to $S_i$ is detected by computing the pixel differences between consecutive frames (as they appear) and by comparing the difference to a predefined threshold. The accuracy of shot boundary detection is not critical to the performance of our method, hence we stick to this simple frame differencing method. 
 
Once we have detected the boundaries of $\mathcal{S}_i$, a standard face detector \cite{dlib09} is employed to detect the faces in each frame in $\mathcal{S}_i$. This frame level face detection can be done in parallel to searching for shot boundaries. The face detector returns the bounding boxes of each face detected in every frame. To build a robust representation of these faces, we use a pretrained CNN, called the FaceNet \cite{facenet}. 
Each face $\mathbf{f}^p$ is forward-passed through the FaceNet to obtain its corresponding $d$ dimensional feature vector $\mathbf{v}^p$. 

To create face tracks, we use a simple yet effective strategy to combine the faces detected in consecutive frames. Let us define two faces detected in two consecutive frames as $\mathbf{f}^p$ and $\mathbf{f}^q$. The overlap $\mathit{a}(\cdot)$ between the two faces is defined as:
\begin{equation}
\mathit{a}(p, q) = \frac{\mathrm{area}(\mathbf{f}^p\cap \mathbf{f}^q)}{\max (\mathrm{area}(\mathbf{f}^p), \mathrm{area}(\mathbf{f}^q) )}*100
\end{equation}
where $\mathrm{area}(\mathbf{f})$ is the area of the rectangular bounding box of $\mathbf{f}$. The squared distance between the feature vectors $\mathbf{v}^p$ and $\mathbf{v}^q$ are defined as $\delta(p,q) = \|\mathbf{v}^p - \mathbf{v}^q\|_2^2$. 
If $\mathit{a}(p, q)> 0.85$ and $\delta(p,q) \leq 1.0$ i.e.,\ if the faces have more than $85$\% overlap and less than $1.0$ feature distance in consecutive frames, they are considered to be of the same person (see Fig.~\ref{fig:FT}). Detected faces that overlap this way in consecutive frames are combined to form a \emph{face track}, and the sequence of features corresponding to each of these faces is defined as a \emph{feature track}.\\

\noindent \textbf{Online face clustering.}
\label{subsec:online}
The next task is to cluster the face tracks as they appear in each shot. For this subtask, we use our recently developed online clustering algorithm \cite{kulshreshtha2018online}. We assume the availability of all face tracks in a single shot at a given time. Our goal is to assign a face track belonging to a person who has appeared earlier to the correct existing cluster, and to form a new cluster for a face track belonging to a person appearing for the first time.  

 Let us consider a shot ${S}_i$ containing $K$ face tracks $\{\mathcal{F}_k\}_{k=1}^{K}$. Each face track $\mathcal{F}_k$ is associated with a feature track $\mathcal{V}_k = \{\mathbf{v}^1_k, \mathbf{v}^2_k, \ldots \mathbf{v}^{{N_k}}_k\}$, where $N_k$ is the number of faces in $\mathcal{F}_k$. 
Also consider that we have already processed previous $(i-1)$ shots and have obtained $L$ clusters corresponding to unique $L$ characters. The clusters are represented by their corresponding cluster centers $\mathcal{C} = \{\mathbf{c}_l\}_{l=1}^L$, where $\mathbf{c}_l \in \mathbb{R}^d$, where $\mathbf{c}_l$ is the feature vector obtained by averaging all features across all face tracks within the $l^{th}$ cluster. Note that the number of clusters and the clusters themselves are dynamic and they evolve as each shot is processed. 
We now define two matrices as follows:
\begin{itemize}

\item A \emph{temporal constraint matrix} $\mathbf{Q} \in \mathbb{R}^{K\times K}$ is defined as
\begin{equation}
\mathbf{Q}(p,q) =
	\begin{cases}
	0 &  \mbox{if $\mathcal{F}_p$ and $\mathcal{F}_q$ overlap in time} \\
	1        & \mathrm{otherwise}
	\end{cases} 
    \label{eq:temp}
\end{equation}
where, $p, q \in \{1, 2, \ldots, K\}$. The matrix $\mathbf{Q}$ enforces a temporal constraint on the face tracks such that if two face tracks have any overlap in time, they are considered to belong to two different characters, and hence, are assigned to different clusters. 
\item A \emph{similarity matrix} $\mathbf{D}\in \mathbb{R}^{L\times K}$ that measures the similarity between a face track (represented by $\mathcal{V}_k$) and a cluster center $c_l$ for a given shot.  
\begin{equation}
\mathbf{D}(l,k) = d(\mathbf{c}_l, \mathcal{V}_k) = 4 - \frac{1}{N_k}\sum_{j=1}^{N_k} \|\mathbf{v}_k^j - \mathbf{c}_l\|_2^2
\label{eq:sim}
\end{equation}
where $l=1,2,\ldots,L$, and $k=1,2,\ldots,K$. The second component is an average squared distance, the maximum value of which is $4$ (since each feature is a unit vector). By subtracting the distance from $4$ we obtain a similarity value between $[0,4]$.
\end{itemize}
\begin{algorithm}[tb]
\caption{Facetrack clustering in a single shot.}
\SetAlgoLined
\DontPrintSemicolon
\SetKwInOut{Initialize}{Initialize}
\KwIn{Face track features in the current shot: $\{\mathcal{V}_k\}_{k=1}^K$, Initial clusters: $\mathcal{C}$}
\KwOut{Updated $\mathcal{C}$}
\Initialize{$\mathbf{ind} = [1,2,\ldots,K]$, $\mathbf{W} =$ all-ones matrix.}\;
    Compute $\mathbf{Q}$, $\mathbf{D}$ using \eqref{eq:temp} and \eqref{eq:sim}\;
  \While{length($\mathbf{ind}) > 0$}{
	\eIf{$\mathcal{C}$ \text{not empty \&\&} $\max_{l,k}( \mathbf{D} \odot \mathbf{W}) \geq \tau$}{
    	$(\hat{l}, \hat{k}) \leftarrow \argmax_{l,k} ( \mathbf{D} \odot \mathbf{W} ) $ \; 
    $k^* \leftarrow \mathbf{ind}[\hat{k}]$\;
Update cluster center $\mathbf{c}_{\hat{j}}$ with $\mathcal{V}_{k^*}$ \; 
    }{
    Add new cluster $(\hat{l}, \hat{k}) \leftarrow ( L+1, 1)$ \; 
    $k^* \leftarrow \mathbf{ind}[\hat{k}]$ \;
 $\mathbf{c}_{new} \leftarrow$ mean($\mathcal{V}_{k^*}$) \;
        $\mathcal{C}$ $\leftarrow$ $\mathcal{C} \cup \mathbf{c}_{new}$ \;
    }
    Recompute $\mathbf{D}$ for $\mathbf{c}_{\hat{l}}$\;
	$\mathbf{W}(\hat{l},:) \leftarrow \mathbf{Q}(\hat{k},:)$ \;
	Delete $\mathbf{D}[:,\hat{k}], \mathbf{W}[:,\hat{k}], \mathbf{Q}[\hat{k},:], \mathbf{Q}[:,\hat{k}], \mathbf{ind}[\hat{k}]$  
}
\label{algo:CT}
\end{algorithm}

Given $\{\mathcal{V}_k\}_{k=1}^K$, our task is to assign them to either one of the $L$ clusters or create new clusters, if required. This is done by simply computing the similarities between $\mathcal{V}_k$ for all $k$ and $\{\mathbf{c}_l\}_{l=1}^L$.
\begin{align}
(\hat{l}, \hat{k}) = \argmax_{l,k} ( \mathbf{D} \odot \mathbf{W})\,
\end{align}
where, $\mathbf{W} \in \mathbb{R}^{L\times K}$ is a weight matrix (initialized with all ones) and $\odot$ denotes element wise product.
If $\max_{l,k}(\mathbf{D} \odot \mathbf{W}) \geq \tau$, where $\tau$ is an user defined threshold $\mathcal{V}_{\hat{k}}$ is assigned to the ${\hat{l}^{th}}$ cluster. Consequently, we update $\mathbf{c}_{\hat{l}}$ by averaging over the existing and the newly added face track. On the other hand, if $\max_{l,k}(\mathbf{D} \odot \mathbf{W}) < \tau$, a new cluster is created assuming a new character has appeared. We add a new cluster $\mathcal{C} \leftarrow \mathcal{C} \cup \mathbf{c}_{new}$. Note that since $\mathbf{W}$ is initialized as a matrix of all ones, it has no effect on the clustering of the first face track. For the subsequent assignments $\mathbf{W}$ is updated to add temporal constraints. 
After $\mathcal{V}_{\hat{k}}$ is assigned to a cluster, we update $\mathbf{D}$ and $\mathbf{W}$ as follows:
\begin{itemize}
	\item Case I: $\mathcal{V}_{\hat{k}}$ is assigned to an existing cluster $\hat{l}$
\begin{equation}
	\mathbf{W}(\hat{l},:) \leftarrow \mathbf{Q}(\hat{k},:) 
\end{equation}
This updated $\mathbf{W}$ will make $\mathbf{D}\odot \mathbf{W}$ zero for all the face tracks having any temporal overlap with $\mathcal{V}_{\hat{k}}$ in the $\hat{l}^{th}$ row.
\begin{equation}
   \mathbf{D}(\hat{l},k) = d(\mathbf{c}_{\hat{l}}, \mathcal{V}_k) \text{for $k \in [1,\vert \mathbf{ind} \vert$]}
\end{equation}
    \item Case II: $\mathcal{V}_{\hat{k}}$ is assigned to a new cluster
\begin{align}
   \hat{l} &= \vert\mathcal{C}\vert+1 \\
   \mathcal{C} &\leftarrow \mathcal{C} \cup \mathbf{c}_{new}\\
   \mathbf{W}(\hat{l},:) &\leftarrow \mathbf{Q}(\hat{k},:)\\
   \mathbf{D}(\hat{l},k) &= d(\mathbf{c}_{new}, \mathcal{V}_k)\, \text{for $k\in[1,\vert \textbf{ind} \vert]$}
   \end{align}
\end{itemize}
where $\mathbf{ind} = [1,2,\cdots K]$. As $\mathcal{V}_{\hat{k}}$ is processed and sent into a cluster, its id is removed i.e.\ $\hat{k}^{th}$ element of $\mathbf{ind}$, $\hat{k}^{th}$ column of $\mathbf{D}$ and $\mathbf{W}$, and $\hat{k}^{th}$ row and column of $\mathbf{Q}$ are removed. 

This process goes on until all tracks in $S_i$ are processed, and then we move to the next shot. We also keep track of the clusters that are updated during each shot. This information is later used to create and update the CIG. Algorithm \ref{algo:CT} summarizes our proposed online face clustering algorithm. 
\subsection{CIG construction}
We now describe the method to construct and update the CIG based on the online face clustering results. Each node in the CIG represents a single cluster corresponding to a character, and each edge captures the interaction between the two characters it connects. In our approach, the CIG is created in parallel to the online face clustering process, where new nodes are added to the CIG and the edge weights are updated after each shot is processed.

We define the relationship between two characters $p$ and $q$ in terms of their temporal co-occurrence in the same or consecutive shots. Considering an adjacency matrix $\mathbf{A}$ the relationship between $p$ and $q$ is formally defined as follows. 
\begin{equation}
\label{eq:charrel}
\begin{split}
\mathbf{A}(p, q) = \sum_i [ \mathbb{I}(p \in \mathcal{S}_{i} \,\&\, q \in \mathcal{S}_{i-1}) + \mathbb{I}(p \in \mathcal{S}_{i} ,\&\, q \in \mathcal{S}_{i}) \\
+ \mathbb{I}(p \in \mathcal{S}_{i} ,\&\, q \in \mathcal{S}_{i+1}) ]
\end{split}
\end{equation}
where $\mathbb{I}(.)$ is the indicator function. 
This count defines the strength of the edge between $p$ and $q$ nodes in the CIG, and is denoted by the element $\mathbf{A}(p,q)$. A diagonal element $\mathbf{A}(p,p)$ denotes the number of times a character $p$ appears in two consecutive shots. To construct and update $\mathbf{A}$ in an online fashion, we begin with an empty $\mathbf{A}$ and keep populating it with new rows and columns (corresponding to newly added nodes and edges) as new shots are processed. The dimension of $\mathbf{A}$ thus increases as new characters are discovered, and consequently, new nodes are added to the CIG. According to our definition of character relationship in eq.~\eqref{eq:charrel}, we need to look for the characters in the shot immediately before and after it. Since we can not peek into the future shot, at shot $S_i$ ($i > 2$), we update $\mathbf{A}$ for $S_{i-1}$. 

Our clustering algorithm yields updated cluster ids $\mathcal{U}_{i-2}$, $\mathcal{U}_{i-1}$, and $\mathcal{U}_{i}$ pertaining to the shots $S_{i-2}, S_{i-1}$, $S_i$. We append $N_{new}^{i-1}$ rows and $N_{new}^{i-1}$ columns to $\mathbf{A}$ (all new elements initialized to 0), where $N_{new}^{i-1}$ is the number of new clusters added during $({i-1})^{th}$ shot. Then $\mathbf{A}(p, q)$ is updated as follows.
\begin{equation}
\begin{split}
\mathbf{A}(p,q) \leftarrow \mathbf{A}(p,q) + \mathbb{I}(p \in \mathcal{U}_{i-1} \& q \in \mathcal{U}_{i-2}) \\
+ \mathbb{I}(p \in \mathcal{U}_{i-1} \& q \in \mathcal{U}_{i-1}) \\
+ \mathbb{I}(p \in \mathcal{U}_{i-1} \& q \in \mathcal{U}_{i})
\end{split}
\label{eq:cig}
\end{equation}

\begin{figure}[tb]
\setlength{\fboxsep}{3pt}%
\setlength{\fboxrule}{1pt}%
\begin{center}
\fbox{\includegraphics[width=0.5\linewidth, trim ={3cm 2.8cm 2.3cm 2cm}, clip]{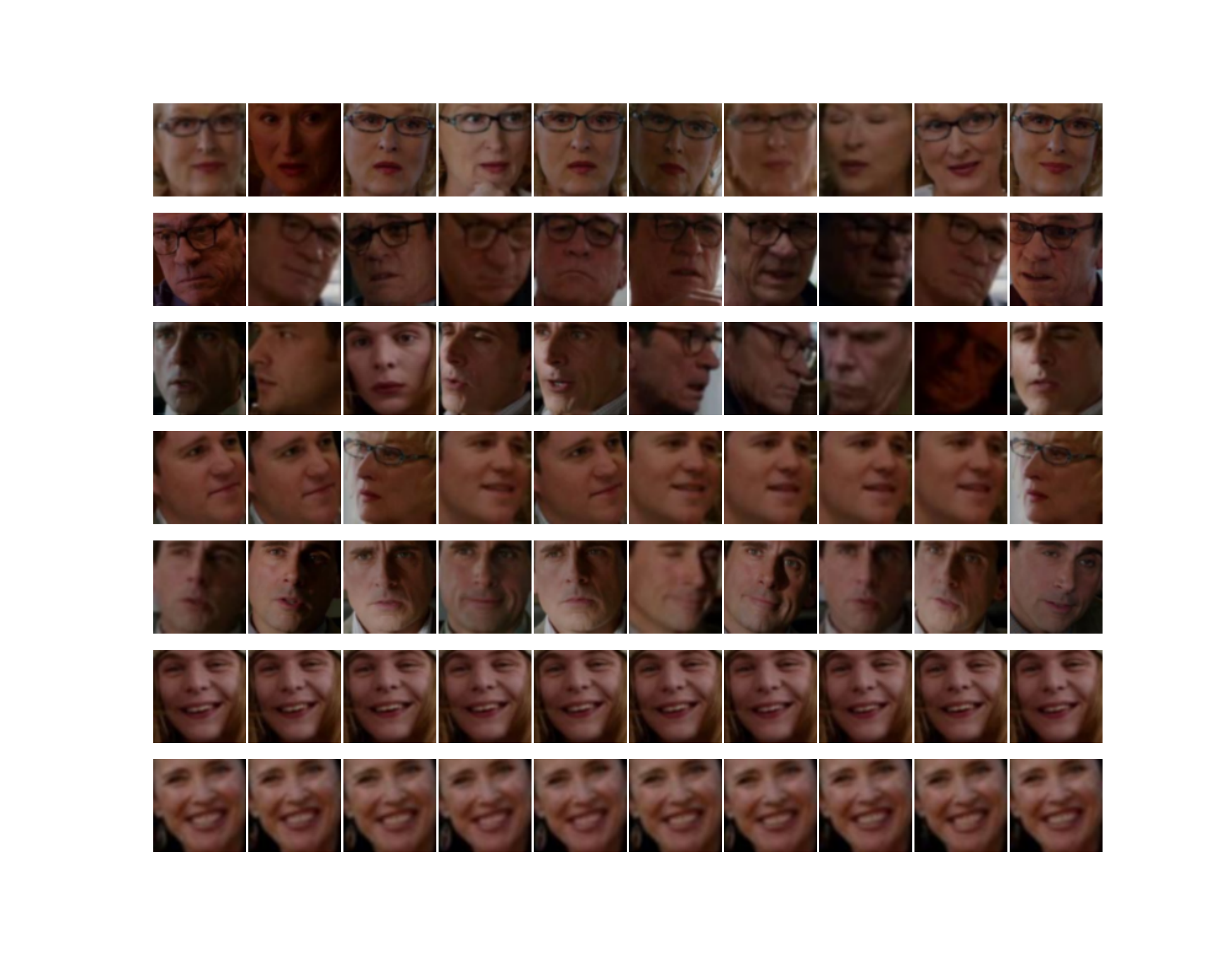}}
\fbox{\includegraphics[width=0.39\linewidth, trim ={4cm 1.1cm 3.3cm 0cm}, clip]{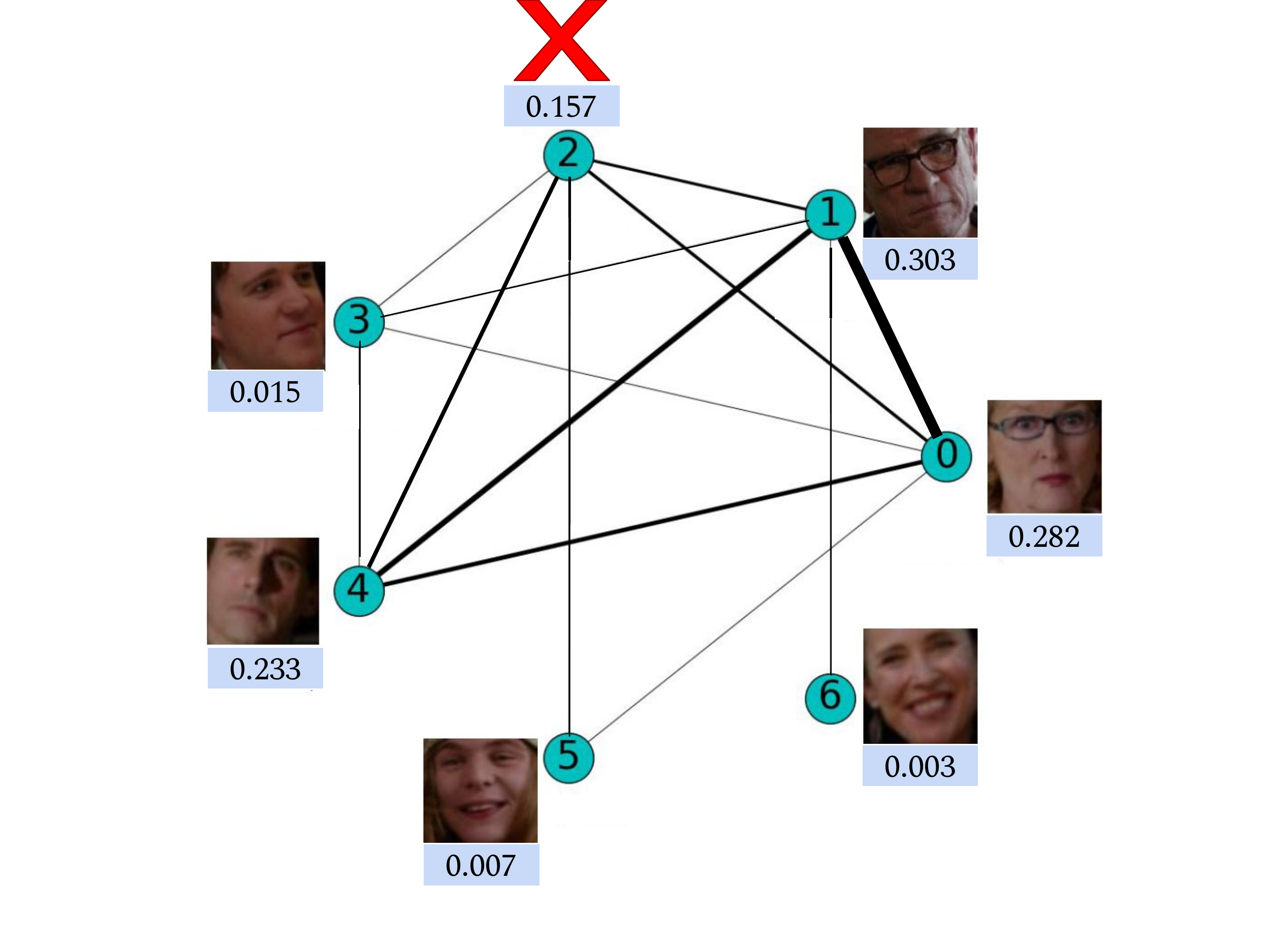}}
\end{center}
\vspace{-3mm}
\caption{Character clusters (left) and the constructed CIG (right) for the movie \textit{Hope Springs}. The CIG shows 7 nodes corresponding to the 7 clusters discovered by our algorithm. The node marked 'X' denotes a noisy cluster. The numbers below each node in the CIG denote the importance ($\sigma(p)$) of the characters.}
\label{fig:CIG_hs}
\end{figure}

\noindent 
{Algorithm \ref{algo:OC}} summarizes the entire process of online clustering and CIG creation as they are performed in parallel. Fig.~\ref{fig:CIG_hs} shows an example of a CIG created using the proposed approach for a movie called \emph{Hope Springs}. The CIG has $6$ pure clusters corresponding to the $6$ characters discovered by our online clustering algorithm and a noisy cluster denoted by 'X'. The edges depict the relationship between the characters where thicker edges denote higher interaction. The numbers represent the character importance scores, later described in Section 4.2 in detail. 
\begin{algorithm}[tb]
\caption{Character interaction graph (CIG) construction via online clustering}
\SetAlgoLined
\DontPrintSemicolon
\SetKwInOut{Initialize}{Initialize}
\KwIn{Movie frames: $I_1$,...$I_{T}$}
\KwOut{}
\Initialize {$i \leftarrow 1$, $T_1 \leftarrow 1$, $\mathcal{C} \leftarrow \{\}$, $\mathbf{A}_{CIG} \leftarrow \phi$} \;
\For{t = $1$ to $T$}{
	\If{shot boundary detected}{
    	$T_{i+1} \leftarrow t$ \;
		Shot $\mathcal{S}_i \leftarrow \{I_{T_i}, I_{T_i+1} 		\ldots I_{T_{i+1}-1}\}$ \;
		Create facetracks [$\mathcal{F}_1, \ldots,  \mathcal{F}_{K}$] from $\mathcal{S}_i$ \;
    	Extract features [$\mathcal{V}_1,		\ldots, \mathcal{V}_{K}$];
		Set of cluster centers $\mathcal{C} = 				\{\mathbf{c}_1, \ldots 					\mathbf{c}_{L}\}$ \;
		$\mathcal{C}, \mathcal{U}_i \leftarrow $ 				\textit{ClusterTracks} ($\mathcal{V}$, $\mathcal{C}$ ) (see Algo.~\ref{algo:CT})\;
    	\If{$i > 2$}{
    		$N_{i-2} \leftarrow$ No. of clusters at $(i-2)$th shot \;
        	$N_{i-1} \leftarrow$ No. of clusters at $(i-1)$th shot \;
     		$\mathbf{A}_{CIG} \leftarrow$ \textbf{\textit{UpdateCIG}} ($\mathcal{U}_{i-2}$, $\mathcal{U}_{i-1}$, $\mathcal{U}_{i}$, $N_{i-2}$, $N_{i-1}$ ) (Using eqn. \ref{eq:cig})\;
        }
        }
    }
\label{algo:OC}
\end{algorithm}
\section{Applications to movie analysis}
\label{sec:application}
In this section, we demonstrate the usefulness of the CIGs for two important movie analysis tasks: (i) Three act segmentation: detecting high level semantic structures in a movie, and (ii) Major character identification. Below, we describe in detail how CIG can facilitate these tasks. 
\subsection{Three act segmentation}
\label{subsec:three_act}
\begin{figure}[tb]
	\includegraphics[width = 0.8\linewidth, trim={0cm 14cm 0cm 0cm}, clip=true]{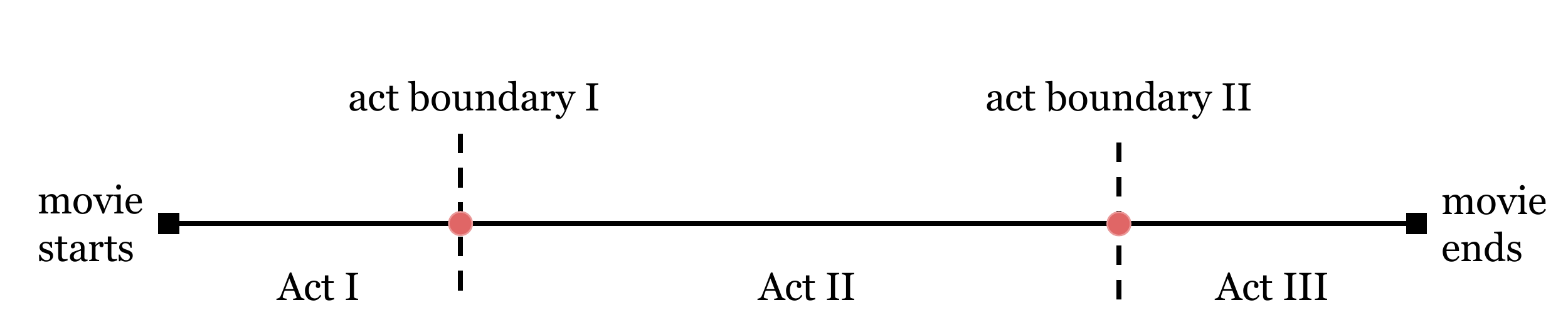}
    \caption{The three-act narrative structure of a movie.}
    \label{fig:3acts}
\end{figure}

Popular films and screenplays are known to follow a well defined storytelling paradigm. The majority of movies consist of three main segments or acts (see Fig.~\ref{fig:3acts}): \emph{Act I} - introduces the main characters and presents a key incident or plot point that drives the story, \emph{Act II} - consists of a series of events including a key event which prepares the audience for the climax, and \emph{Act III} - includes the climax and the resolution of the story \cite{field2007screenplay, mckee1997substance}. Discovering these high-level semantic units automatically can help in movie summarization and detection of the key events \cite{movieNarrativesTGuha}.  

Our objective is to segment a movie into its three acts by detecting the two act boundaries as shown in Fig.~\ref{fig:3acts}. Consider the CIGs $\mathbf{A}_{S_{i-1}}$ and $\mathbf{A}_{S_{i}}$ obtained at shots $S_{(i-1)}$ and $S_i$ respectively. The difference between two CIGs is computed using \emph{graph edit distance} (GED) as follows:
$$\text{GED}(\mathbf{A}_{\mathcal{S}_{(i-1)}} ,\mathbf{A}_{\mathcal{S}_{i}}) = \Delta \eta + \Delta e$$
where, $\Delta \eta$ is the number of new nodes added to $\mathbf{A}_{S_{i}}$, and $\Delta e$ is the number of edges that are modified to obtain $\mathbf{A}_{S_{i}}$ from $\mathbf{A}_{S_{i-1}}$. 

Using this measure, we compute how the CIG for a given movie changes over time between consecutive shots. A window of length $T_w$ is used to sum all the GED scores within the window to incorporate a longer context and get a measure of overall interaction around each shot. Let this CIG difference be denoted as $y^{ged}$, where $y^{ged}_i$ represents the changes in interaction around shot $S_i$. 
We detect act boundary I as follows
\begin{equation}
t_{b1} = \frac{\sum_{i\in \mathcal{B}_1} t_iy^{ged}_i}{\sum_{i\in \mathcal{B}_1} t_i}
\end{equation}
where, $t_i$ is the time at the center of $S_i$, and $B$ is a predefined interval. This interval  $\mathcal{B}$ is chosen leveraging information from film grammar \cite{field2007screenplay}, which suggests that act boundary I lies within $25$ to $30$ minutes from the start of the movie. We thus set $\mathcal{B}$ to have all the shots between an interval of $22$ to $40$ minutes from the start of the movie. The act boundary II, $t_{b2}$ is detected in a similar fashion with an interval $\mathcal{B}_2$ being $14$ to $34$ minute before the end of the movie.
\subsection{Major character identification}
\label{subsec:char_imp}
Another important task in movie analysis is to identify its major characters. Past work on major character discovery using character networks usually rely on betweenness, centrality and sum of the edge weights \cite{ramakrishna2017linguistic, cocharnet, rolenet}.
We compute a new measure called the \emph{eigenvector centrality} for each character in our CIG. 

The eigenvector centrality, $e_p$ of a character (node) $p$ measures the influence the node $p$ has on the CIG, and is defined as follows:
\begin{equation}
e_p = \frac{1}{\zeta}\sum_{q=1}^{\vert\mathcal{C}\vert} e_q \mathbf{A}(p,q)  
\end{equation}
where, $\zeta$ is the largest eigenvalue of $\mathbf{A}$, and $\mathbf{A}(p,q)$ denotes the weight of the edge between nodes $p$ and $q$. We then define a \emph{node importance measure} $\sigma(p)$ for node $p$ as follows:
\begin{align*}
\sigma(p) = \frac{e_p}{\sum_q e_q}.
\end{align*}
It is easy to see that the higher the value of $\sigma(p)$, the more important is the node (character). We use the values of $\sigma(p)$ to rank the movie characters in terms of their importance in the movie. For example, Fig.~\ref{fig:CIG_hs} shows these node importance measures for the characters in a movie.
\section{Performance evaluation}
\label{sec:evaluation}
In this section, we present results and performance comparisons for the different components of our proposed method. First, we present results on the performance of the online face clustering algorithm as it is a critical component of the CIG construction algorithm, and its accuracy determines the quality of the CIG. Direct evaluation of a CIG is not very meaningful, as CIGs may have different characteristics by construction. Hence, we evaluate the usefulness of the CIGs vis two movie analysis tasks - act segmentation and major character discovery.

\subsection{Evaluating clustering performance}
\noindent\textbf{\emph{Databases:}} We use two databases that are commonly used to benchmark face clustering algorithms: (i) \textit{Buffy} database (BF2006) \cite{buffy2006, constrainedclustering} containing $229$ face tracks of $6$ characters ($17,337$ faces altogether) extracted from the episode 2, season 5 of the TV series \textit{Buffy - the Vampire Slayer}. The database includes the frame number, bounding box coordinates, track ids, and the character names for each face. (ii) \textit{Notting Hill} database (NH2016) \cite{baoyuan16} that contains $277$ face tracks of $7$ characters ($19,278$ faces altogether) from the movie \emph{Notting Hill}. It contains the frame numbers, bounding boxes, track ids, features and character names for each face in the database. \\

\vspace{-3mm}\noindent\textbf{\emph{Experimental details:}} For each video, we obtain shot boundaries, create face tracks, extract deep features and cluster the faces using our proposed algorithm (see Algorithm \ref{algo:CT}). We use the FaceNet \cite{facenet} to extract features from each face in a face track. The value of the threshold parameter $\tau$ is set to $2.80$ and $ 2.85$ for the BF2006 and the NH2016 database. For BF2006, we get a cluster for each of the 6 characters, and for NH2016, we get a cluster for 6 out of the 7 characters.\\

\vspace{-3mm}\noindent\textbf{\emph{Comparison with existing methods:}} We compare with two baselines ((i) Gaussian mixture model (GMM) with FaceNet features, (ii) Kmeans with FaceNet features) several state-of-the-art face clustering methods: (i) ULDML \cite{uldml}, (ii) a recently proposed constrained clustering method - the coupled HMRF (cHMRF) \cite{baoyuan16}, and (iii) an aggregated CNN feature-based clustering (aCNN) \cite{simplefc}.
Performances of all the methods are compared in Table \ref{tbl:all} in terms of clustering accuracy (expressed in \%) which compares the predicted cluster labels with the ground truth labels. Note that all the methods in Table \ref{tbl:all} are offline methods, where the entire data, information about the face tracks and the cluster counts are provided as an input to the algorithms. For the online method, however, no information about the face tracks or cluster counts are available.
The performance of our algorithm on the BF2006 database is superior to that of cHMRF and ULDML, and is comparable to Kmeans. On the NH2016 database, our algorithm outperforms all its offline counterparts, achieving a clustering accuracy of $93.8\%$.
\begin{table}[t]
\begin{center}
\renewcommand{\arraystretch}{1.2}
\caption{Comparison with the state-of-the-art (offline) clustering methods in terms of clustering accuracy (\%)} 
\label{tbl:all}
\begin{tabular}{|l c c|}
  \hline
\bf Method   & \bf BF2006 & \bf NH2016 \\
  \hline \hline
ULDML
\cite{uldml} & 49.29 & 43.82 \\
cHMRF
\cite{baoyuan16} & $61.87$ & $47.94$\\
Luo et al. \cite{jointface} & $92.13$ & - \\
FaceNet + aCNN
\cite{simplefc} & $\mathbf{89.90}$ & $90.17$ \\
FaceNet + GMM & $84.21$  & $73.46$ \\
FaceNet + Kmeans & $82.92$ & $71.66$\\
\bf Ours & $82.12$ &   $\mathbf{93.84}$ \\
  \hline
\end{tabular}
\end{center}
\end{table}
\begin{table}[tb]
\centering\caption{Comparison with the existing online clustering method}
\label{tbl_comp}
\renewcommand{\arraystretch}{1.2}
\begin{tabular}{|l|c c| c c|}
\hline 
\bf  & \multicolumn{2}{c}{\bf BF2006} & \multicolumn{2}{|c|}{\bf NH2016}\\  
&  TCCRP\cite{bayesianentity} & Ours &  TCCRP\cite{bayesianentity} & Ours \\
 \hline \hline
Homogeneity & $\bf 0.93$ & $0.68$ & $\bf 0.92$ & $0.88$\\
Completeness & $0.44$ & $\bf 0.69$ & $0.44$ & $\bf 0.89$\\
\emph{V} measure & $0.60$ & $\bf 0.68$ & $0.58$	&$\bf 0.89$\\
Clusters formed & $57$ & $7$ & $61$ & $7$\\ \hline
\end{tabular}
\end{table}

We next compare with the only existing online face clustering algorithm TCCRP \cite{bayesianentity}. We combine TCCRP with FaceNet features, and used a tracklet length of $10$. 
Comparison is made in terms of \emph{homogeneity} score, \emph{completeness} score and their harmonic-mean i.e., the \emph{V measure} (see Table \ref{tbl_comp}). 
Table \ref{tbl_comp} shows that TCCRP has higher cluster homogeneity, but this is achieved at the cost of over-clustering (note the large number of clusters created by TCCRP) and thereby degrading completeness. Our method achieves significantly higher completeness and V measure while discovering a more accurate number of clusters.
\subsection{Evaluating CIGs for act segmentation}
\noindent\textbf{\emph{Database:}} We use a database of six full length Hollywood movies: \textit{Good Deeds, Hope Springs, Joyful Noise, Resident Evil, Step Up Revolution, The Vow}. These movies are known to have a well defined three act structure \cite{movieNarrativesTGuha}. The labels for the act boundaries for the movies were annotated by three film experts. Each expert independently marked the act boundaries for each movie, and then decided on a final time stamp (at the precision level of seconds) through discussions \cite{movieNarrativesTGuha}.\\ 

\vspace{-3mm}\noindent\textbf{\emph{Experimental Set up:}} We run the DLib face detector \cite{dlib09} on each frame of the movie,  and create face tracks. For removing the false detections and very small tracks we set the feature-distance threshold $\delta$ for track creation at $1.0$, and the spacial overlap threshold $\alpha$ is set to $0.95$. The online clustering threshold $\tau$ is set to $3.0$ for all the movies. Also, the face tracks of length less than 15 are discarded.\\
\begin{table}[tb]
\renewcommand{\arraystretch}{1.2}
\caption{Results on act boundary detection: performance measured in terms of the distance from human expert annotated labels (in seconds).} 
\label{tab:3acts}
\scriptsize
\centering
\setlength{\fboxsep}{0pt}%
\setlength{\fboxrule}{2pt}%
\begin{tabular}{|l|c c c c c|c|}
  \hline
\bf Movie   & {\bf Video}\cite{movieNarrativesTGuha} & {\bf Music} \cite{movieNarrativesTGuha} & {\bf Text}\cite{movieNarrativesTGuha} & {\bf Multimodal}\cite{movieNarrativesTGuha} & {\bf Baseline} & \bf Ours\\
  \hline \hline
  \multicolumn{7}{c}{\emph{Act boundary I}}\\
\hline 
\textit{Good Deeds}   &  $309$   &  $36$   &  $622$   &  $34$   &  $329$   &  $\bf 11$ \\  
\textit{Hope Springs}   &  $\bf 13$   &  $34$   &  $31$   &  $31$   &  $620$   &  $99$   \\  
\textit{Joyful Noise}  &  $98$   &  $143$   &  $452$   &  $19$   &  $28$   &  $\bf 17$   \\  
\textit{Resident Evil}  &  $48$   &  $134$   &  $275$   &  $56$   &  $351$   &  $\bf 27$   \\  
\textit{Step Up Rev.}  &  $\bf 96$   &  $346$   &  $277$   &  $265$   &  $578$   &  $289$   \\  
\textit{The Vow} &  $198$   &  $125$  &  $\bf 25$   &  $316$   &  $73$   &  $58$   \\ 

\textbf{\emph{All movies}} & $127.0$ & $136.3$ & $280.3$ & $120.1$ & $329.8$ & $\bf 83.5$ \\
\hline
\multicolumn{7}{c}{\emph{Act boundary II}}\\
  \hline
\textit{Good Deeds}   &  $612$   &  $309$   &  $904$   &  $300$   &  $\bf 207$   &  $252$   \\
\textit{Hope Springs}   &  $210$   &  $1672$   &  $828$   &  $205$   &  $393$   &  $\bf 85$ \\
\textit{Joyful Noise}   &  $528$   &  $320$  &  $263$   &  \bf $146$   &  $163$   &  $177$  \\
\textit{Resident Evil}   &  $836$   &  $575$   &  $702$   &  $575$  &  $751$   &  $\bf 52$  \\
\textit{Step Up Rev.}   &  $522$  &  $507$   &  $480$   &  $466$   &  $407$   &  $\bf 323$  \\
\textit{The Vow}     &  $413$   &  $2053$   &  $\bf{43}$   &  $\bf{43}$   &  $160$   &  $294$   \\  
\textbf{\emph{All movies}} & $520.1$ & $906.0$ & $536.6$ & $289.1$ & $346.8$ & $\bf{197.1}$\\
\hline
\end{tabular}
\end{table}
\begin{figure} [t]
\centering
\includegraphics[width=1.0\textwidth, height = 0.4\textwidth, trim ={0cm 6cm 0cm 3cm}, clip]{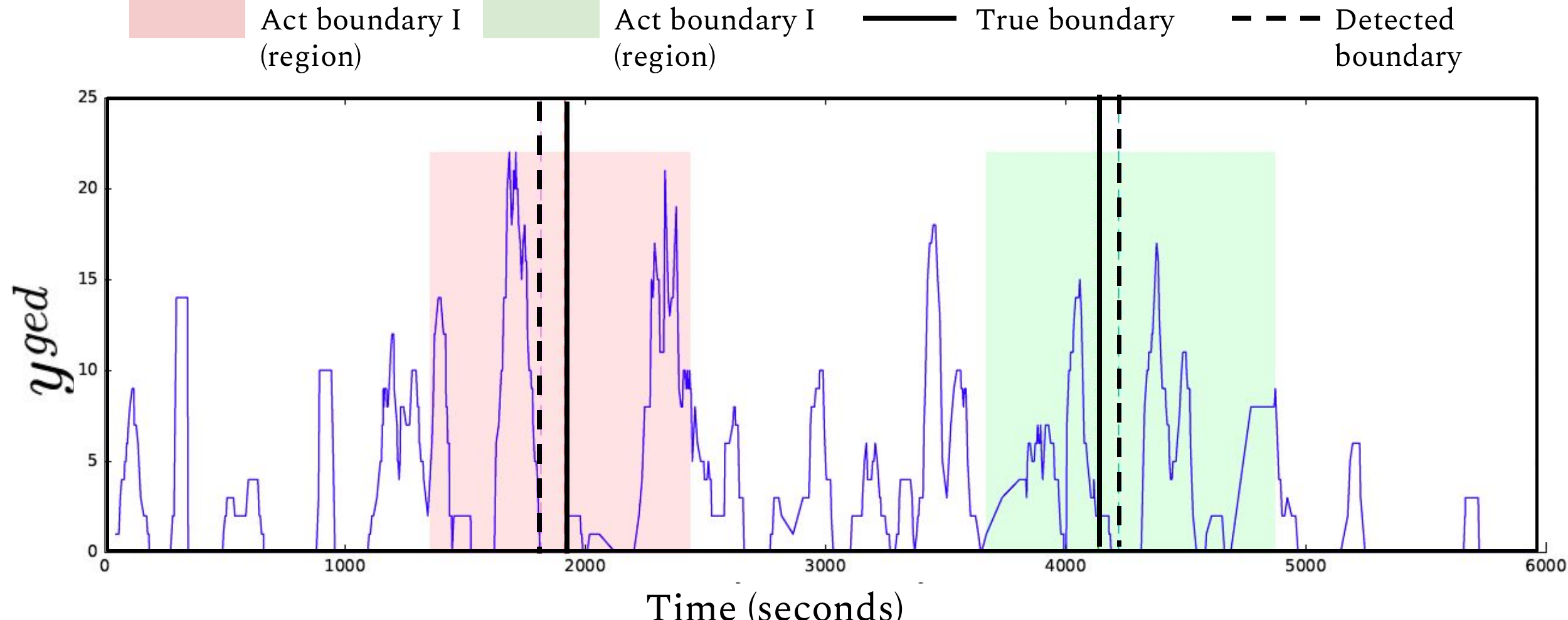}
\caption{Act boundary detection result for the movie \textit{Hope Springs}.}
\label{fig:AB}
\end{figure}

\vspace{-3mm}\noindent\textbf{\emph{Results and discussion:}} 
We detect the two act boundaries (see Fig.~\ref{fig:3acts}) in each movie using the CIGs as described in Section \ref{subsec:three_act} and compute the error in terms of the distance (in seconds) from the expert annotations.  The parameter value of $T_w$ is set to 60s. We compare the performance of our CIG-based approach with that of an existing multimodal approach proposed in an earlier work \cite{movieNarrativesTGuha}. We also create a simple baseline for comparison. The \emph{baseline} sets the first act boundary at 25th minute mark of the movie, and the second act boundary is at 25th minute mark from the end of the movie. Table \ref{tab:3acts} present all the results of act boundary detection for the proposed method along with the baseline and the multimodal approach \cite{movieNarrativesTGuha}. Our CIG-based approach performs the best in terms of overall error, even though it uses information from only the visual stream. We also note that detecting act boundary II is more difficult as it has higher variability across movies. Figure \ref{fig:AB} presents an example of CIG distance plot and the detected act boundaries for the movie \textit{Hope Springs}. 
\begin{table}[tb]
\renewcommand{\arraystretch}{1.2}
\vspace{-2mm}
\begin{center}
\caption{Face track statistics and clusters formed using online face clustering}
\label{tbl:noisy}
\scriptsize
\begin{tabular}{|l| c c c c | c c  c|}
\hline
\multirow{2}{*}{\bf Movie} & \multirow{ 2}{*}{\bf Faces} & \multirow{ 2}{*}{\bf Shots} & \multirow{ 2}{*}{\bf Tracks} & {\bf Avg track} & {\bf Noisy} & {\bf Char} & {\bf Total}\\
& & & & {\bf length} & {\bf clusters} & {\bf clusters} & {\bf }\\
\hline \hline
\textit{Good Deeds} & $38766$ & $910$ & $1078$ & $35.96$ & $2$ & $7$ & $9$ \\  
\textit{Hope Springs} & $22663$ & $1359$ & $669$ & $33.87$ & $1$ & $6$ & $7$  \\  
\textit{Joyful Noise} & $38703$ &  $2023$ & $1205$ & $32.11$ & $7$ & $14$ & $21$ \\  
\textit{Resident Evil} & $17962$ & $2520$ & $537$ & $33.45$ & $4$ & $8$ & $12$ \\  
\textit{Step Up Rev.} & $21986$ & $2790$ & $733$ & $29.99$ & $6$ & $13$ & $19$ \\  
\textit{The Vow} & $43860$ & $1419$ & $1161$ & $37.77$ & $3$ & $5$ & $8$  \\  
\hline
\end{tabular}
\end{center}
\end{table}
\subsection{Evaluating CIGs for major character identification}
For this task, we use the same six movies as the three act segmentation task described in the previous section. The experimental settings remain the same.\\

\vspace{-2mm}\noindent\textbf{\emph{Results and discussion:}} We first run our online face clustering algorithm on each movie. Some of the clusters thus obtained may be noisy i.e., they may contain faces from multiple characters. Such noisy clusters are formed due to (i) the presence of the minor characters in movies who do not appear on-screen long enough, and (ii) some wrongly clustered faces of major characters. Since the ground truth face labels are not available for the movies, we sought manual validation of the clusters to evaluate the performance of our method. Two human annotators labeled all the clusters formed for each movie, and identified each cluster as either a valid \emph{character cluster} or a \emph{noisy cluster}. The results are presented in Table \ref{tbl:noisy}. 

After the clusters are formed, we compute $\sigma(p)$ for each cluster of any given movie, and identify the top 5 clusters (using the $\sigma(p)$ values) as the five major characters in the movie. To validate the results, we again seek manual evaluation. Two human annotators watched each movie, and based on the internet movie database (IMDb)\footnote{www.imdb.com/} major cast list and the storyline, identified the top 5 characters in each movie. Table \ref{tbl:mc} presents the corresponding results, where 'X' denotes a noisy cluster. The results show that the top two characters are always retrieved correctly, and for most of the cases, our CIG-based approach is able to retrieve four out of the top five characters.   
\begin{table}[t]
\renewcommand{\arraystretch}{1.3}
\begin{center}
\caption{Major character identification results on six full-length movies (`X' denotes noisy cluster).} 
\label{tbl:mc}
\scriptsize
\begin{tabular}{|p{1.6cm}|p{1cm}|c c c c c|c|}
\hline 
\bf Movie & \multicolumn{6}{c|}{\bf Characters in the decreasing order of importance} & \bf Precision \\
\hline \hline
\multirow{2}{*}{\emph{Good Deeds}} & Ours & Lindsey & Wesley& X & Walter& X & \multirow{2}{*}{$0.6$} \\
 & True & Wesley & Lindsey & Natalie & Walter & John &\\
\hline 
\multirow{2}{*}{\emph{Hope Springs}} & Ours & Arnold & Kay & Bernie & X & Brad & \multirow{2}{*}{$0.8$}\\
 & True & Arnold & Kay & Bernie & Eileen & Brad &\\
\hline 
\multirow{2}{*}{\emph{Joyful Noise}} & Ours & Rose & Olivia & Earla & X & Randy & \multirow{2}{*}{$0.8$}\\
 & True & Rose & Olivia & Sparrow & Earla & Randy &\\
\hline
 \multirow{2}{*}{\emph{Resident Evil}} & Ours & Alice & X & Ada & Wesker & Rain & \multirow{2}{*}{$0.8$}\\
 & True & Alice & Ada & Rain & Wesker & Jill &\\
\hline 
  \multirow{2}{*}{\emph{Step Up Rev}} & Ours & Emily & Sean & Eddy & Penelope & William & \multirow{2}{*}{$0.8$}\\
 & True & Emily & Sean & Eddy & William & Jason &\\
\hline
  \multirow{2}{*}{\emph{The Vow}} & Ours & Paige & Leo & X & X & Jeremy & \multirow{2}{*}{$0.6$}\\
 & True & Paige & Leo & Jeremy & Bill & Rita & \\
\hline
\end{tabular}
\end{center}
\end{table}
\section{Conclusion}
\label{sec:conclusion}
We proposed an unsupervised approach to building a \emph{dynamic} character network of movie characters through online face clustering. This is significantly different from the existing body of work that builds a single, static character network using supervision from text, meta-data, or human annotation. 
We demonstrated that the dynamic CIGs can successfully detect high-level semantic structures (acts) in movie narratives, and can also identify the major characters in a movie with high precision. Apart from the applications presented in this paper, dynamic CIGs are also expected to be useful for extracting character-level analytics, movie summarization, indexing and navigation.

Future work will be directed towards expanding the database used for validation, and leveraging the subtitle and audio information available for the movies to achieve better clustering accuracy. A scheme of splitting and fusing the clusters formed online can be useful to improve the quality of clusters, and in turn, can improve the quality of the CIG. 

\bibliographystyle{spmpsci}
\bibliography{main.bib}
\end{document}